\documentclass[sigconf]{acmart}
\renewcommand\footnotetextcopyrightpermission[1]{} 
\usepackage{bbm}
\usepackage{booktabs}

\newcommand{\Xb}{\mathbf{X}}
\newcommand{\M}{\mathbf{M}}
\newcommand{\xb}{\mathbf{x}}
\newcommand{\Yb}{\mathbf{Y}}
\newcommand{\yb}{\mathbf{y}}

\newcommand{\Wb}{\mathbf{W}}
\newcommand{\w}{\textbf{$\boldsymbol{\omega}$}}

\begin{document}

\copyrightyear{2018} 
\acmYear{2018} 
\setcopyright{usgovmixed}
\acmConference[KDD '18]{The 24th ACM SIGKDD International Conference on Knowledge Discovery \& Data Mining}{August 19--23, 2018}{London, United Kingdom}
\acmBooktitle{KDD '18: The 24th ACM SIGKDD International Conference on Knowledge Discovery \& Data Mining, August 19--23, 2018, London, United Kingdom}
\acmPrice{15.00}
\acmDOI{10.1145/3219819.3219996}
\acmISBN{978-1-4503-5552-0/18/08}

\fancyhead{}

\title{Quantifying Uncertainty in Discrete-Continuous and Skewed Data with Bayesian Deep Learning}

\author{Thomas Vandal}
\affiliation{%
  \institution{Northeastern University, Civil and Environmental Engineering}
  \streetaddress{360 Huntington Ave.}
  \city{Boston} 
  \state{MA} 
  \postcode{02141}
}
\email{vandal.t@husky.neu.edu}

\author{Evan Kodra}
\affiliation{%
  \institution{risQ Inc.}
  \streetaddress{404 Broadway}
  \city{Cambridge} 
  \state{MA} 
  \postcode{02139}
}
\email{evan.kodra@risq.io}

\author{Jennifer Dy}
\affiliation{%
  \institution{Northeastern University, Electrical and Computer Engineering}
  \streetaddress{360 Huntington Ave.}
  \city{Boston} 
  \state{MA} 
  \postcode{02141}
}
\email{j.dy@neu.edu}

\author{Sangram Ganguly}
\affiliation{%
  \institution{Bay Area Environmental Research Institute / NASA Ames Research Center}
  \city{Moffett Field} 
  \state{CA} 
  \postcode{94035}
}
\email{sangram.ganguly@nasa.gov}

\author{Ramakrishna Nemani}
\affiliation{%
  \institution{NASA Advanced Supercomputing Division/ NASA Ames Research Center}
  \city{Moffett Field} 
  \state{CA} 
  \postcode{94035}
}
\email{rama.nemani@nasa.gov}

\author{Auroop R Ganguly}
\affiliation{%
  \institution{Northeastern University, Civil and Environmental Engineering}
  \streetaddress{360 Huntington Ave.}
  \city{Boston} 
  \state{MA} 
  \postcode{02141}
}
\email{a.ganguly@neu.edu}

\renewcommand{\shortauthors}{T. Vandal et al.}

\begin{abstract}
Deep Learning (DL) methods have been transforming computer vision with innovative adaptations to other domains including climate change. For DL to pervade Science and Engineering (S\&E) applications where risk management is a core component, well-characterized uncertainty estimates must accompany predictions. However, S\&E observations and model-simulations often follow heavily skewed distributions and are not well modeled with DL approaches, since they usually optimize a Gaussian, or Euclidean, likelihood loss. Recent developments in Bayesian Deep Learning (BDL), which attempts to capture uncertainties from noisy observations, aleatoric, and from unknown model parameters, epistemic, provide us a foundation. Here we present a discrete-continuous BDL model with Gaussian and lognormal likelihoods for uncertainty quantification (UQ). We demonstrate the approach by developing UQ estimates on ``DeepSD'', a super-resolution based DL model for Statistical Downscaling (SD) in climate applied to precipitation, which follows an extremely skewed distribution. We find that the discrete-continuous models outperform a basic Gaussian distribution in terms of predictive accuracy and uncertainty calibration. Furthermore, we find that the lognormal distribution, which can handle skewed distributions, produces quality uncertainty estimates at the extremes. Such results may be important across S\&E, as well as other domains such as finance and economics, where extremes are often of significant interest. Furthermore, to our knowledge, this is the first UQ model in SD where both aleatoric and epistemic uncertainties are characterized. 
\end{abstract}



\maketitle

\section{Introduction}



\begin{figure*}[t]
    \centering
    \includegraphics[width=\textwidth]{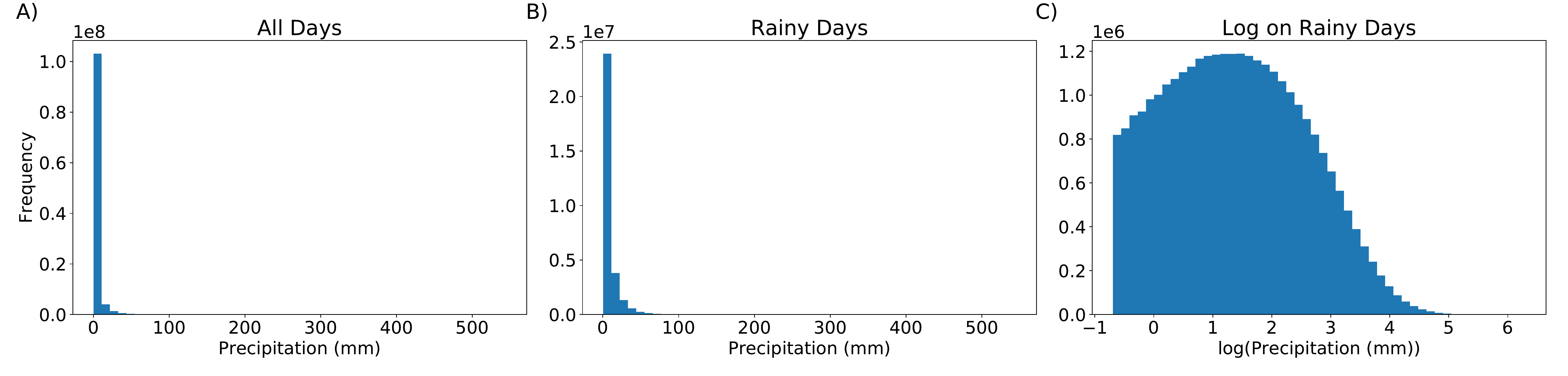}
    \caption{Histogram of daily precipitation on the Contiguous United States from 2006 to 2015. A) All precipitation data points. B) Precipitation distribution on rainy days only.  C) Log distribution of precipitation on rainy days.}
    \label{fig:precip-dist}
\end{figure*}

Science and Engineering (S\&E) applications are beginning to leverage the recent advancements in artificial intelligence through deep learning. In climate applications, deep learning is being used to make high-resolution climate projections~\cite{vandal2017deepsd} and detect tropical cyclones and atmospheric rivers~\cite{racah2017extremeweather}. Remote sensing models such as DeepSAT~\cite{basu2015deepsat}, a satellite image classification framework, also leverage computer vision technologies. Physicists are using deep learning for detecting particles in high energy physics~\cite{baldi2014searching} and in transportation deep learning has aided in traffic flow prediction~\cite{lv2015traffic} and modeling network congestion~\cite{ma2015large}. Scientists have even used convolutional neural networks to  approximate the Navier-Stokes equations of unsteady fluid forces~\cite{miyanawala2017efficient}. However, for many of these applications, the underlying data follow non-normal and discrete-continuous distributions. For example, when modeling precipitation, we see most days have no precipitation at all with heavily skewed amounts on the rainy days, as shown in Figure~\ref{fig:precip-dist}. Furthermore, climate is a complex nonlinear dynamical system, while precipitation processes in particular exhibit extreme space-time variability as well as thresholds and intermittence, thus precipitation data cannot be assumed to be Gaussian. Hence, for deep learning to be harnessed to it's potential in S\&E applications, our models must be resilient to non-normal and discrete-continuous distributions.

Uncertainty quantification is another requirement for wide adoption of deep learning in S\&E, particularly for risk management decisions. Twenty years ago, Jaeger et al. stated, ``uncertainties in climate change are so pervasive and far reaching that the tools for handling uncertainty provided by decision analysis are no longer sufficient~\cite{jaeger1998decision}.'' As expected, uncertainty has been a particular interest of climate and computer scientists to inform social and infrastructure adaptation to increasing weather extremes and natural disasters~\cite{katz2002techniques,lobell2008prioritizing}. For example, Kay et al. studied six different sources of uncertainty of climate change impacts on a flood frequency model~\cite{kay2009comparison}. These uncertainties included future greenhouse gas scenarios, global climate models (GCMs) structure and parameters, downscaling GCMs, and hydrological model structure and parameters. Hence, quantifying the uncertainty from each of these processes is critical for understanding the system's uncertainty.  This provides us with the problem of quantifying uncertainty in discrete-continuous and non-normal distributions.

Recent work in Bayesian Deep Learning (BDL) provides a foundation for modeling uncertainty in deep networks which may be applicable to many S\&E applications~\cite{gal2016uncertainty,gal2017concrete,kendall2017b,zhu2018bayesian}. The simplicity of implementing BDL on an already defined deep neural network makes it an attractive approach. With a well-defined likelihood function, BDL is able to capture both aleatoric and epistemic uncertainty~\cite{kendall2017b}.  \textit{Epistemic} uncertainty comes from noise in the model's parameters which can be reduced by increasing the dataset size. On the other side, \textit{Aleatoric} uncertainty accounts for the noise in the observed data, resulting in uncertainty which cannot be reduced. Examples of aleatoric uncertainty are measurement error and sensor malfunctions. Aleatoric uncertainty can either be homoscedastic, constant uncertainty for different inputs, or heteroscedastic, uncertainty depending on the input. Heteroscedastic is especially important in skewed distributions, where the tails often contain orders of magnitude increased variability. Variants of these methods have already been successfully applied to applications such as scene understanding~\cite{kendall2015bayesian} and medical image segmentation~\cite{wang2018interactive}.  

While BDL has been applied to few domains, these models generally assume a Gaussian probability distribution on the prediction. However, as we discussed in S\&E applications, such an assumption may fail to hold. This motivates us to extend BDL further to aperiodic non-normal distributions by defining alternative density functions based on domain understanding. In particular, we focus on a precipitation estimation problem called statistical downscaling, which we will discuss in Section 2. In section 3, we review ``DeepSD'', our statistical downscaling method~\cite{vandal2017deepsd}, and Bayesian Deep Learning Concepts. In section 4, we present two BDL discrete-continuous (DC) likelihood models, using Gaussian and lognormal distributions, to model categorical and continuous data. Following in Section 5, we compare predictive accuracy and uncertainty calibration in statistical downscaling. Lastly, Section 6 summarizes results and discusses future research directions. 

\subsection{Key Contributions}

\begin{enumerate}
    \item A discrete-continuous bayesian deep learning model is presented for uncertainty quantification in science and engineering.
    \item We show that a discrete-continuous model with a lognormal likelihood can model fat-tailed skewed distributions, which occur often in science and engineering applications.
    \item The first model to capture heteroscedastic, and epistemic, uncertainties in statistical downscaling is presented. 
\end{enumerate}

\section{Precipitation Estimation} \label{sec:data}

\begin{figure}
    \centering
    \includegraphics[width=0.5\textwidth]{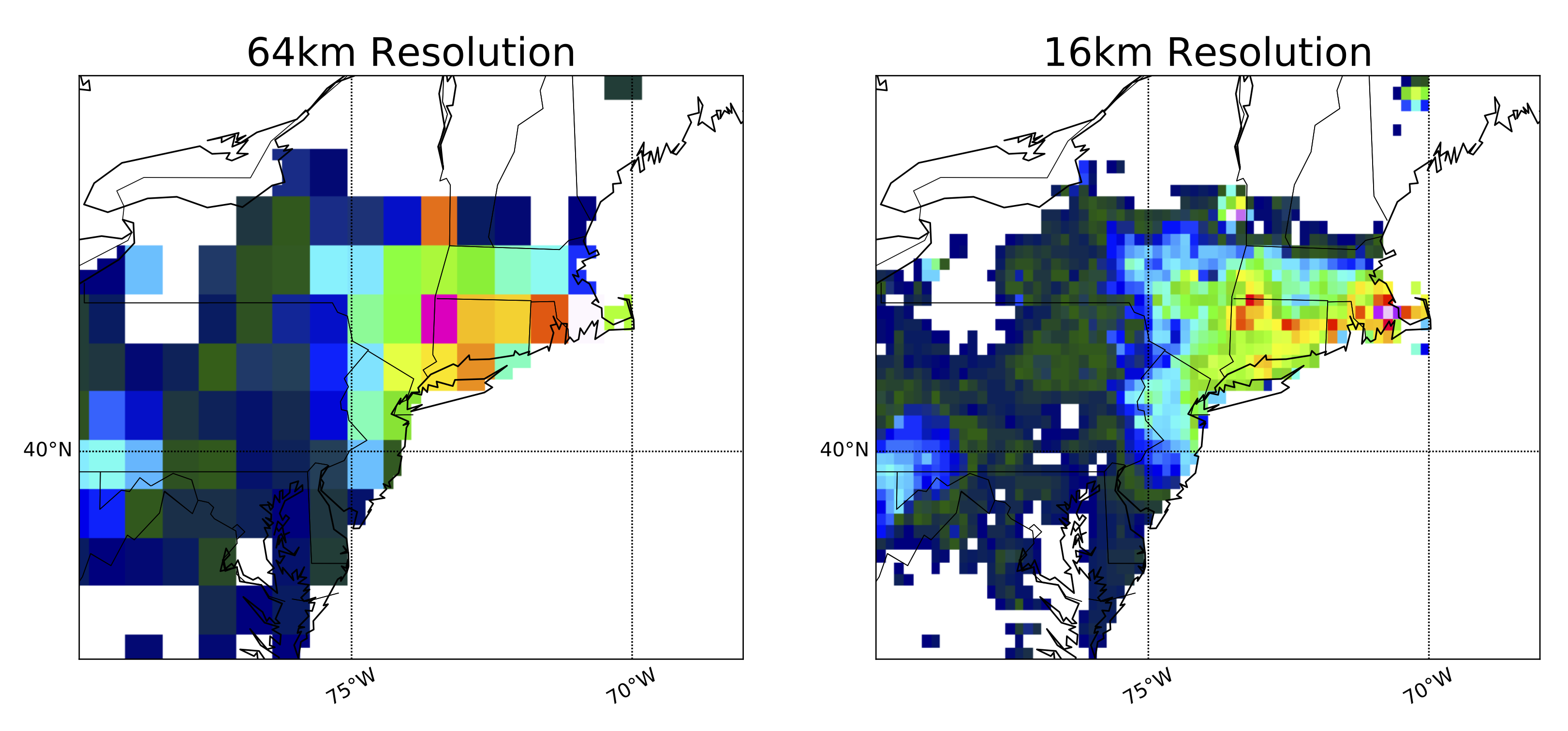}
    \caption{Prism Observed Precipitation: Left) Low resolution at 64km. Right) High resolution at 16km.}
    \label{fig:lr-hr}
\end{figure}

~\subsection{Statistical Downscaling}
Downscaling, either statistical or dynamical, is a widely used process for producing high-resolution projections from coarse global climate models (GCMs)~\cite{hewitson1996climate,fowler2007linking,maraun2010precipitation}. Dynamical downscaling, often referred to as regional climate models, are physics based numerical models encoding localized sub-grid processes within GCM boundary conditions to generate high-resolution projections. Similar to GCMs, dynamical downscaling are computational expensive and simply cannot scale to ensemble modeling. Statistical downscaling is a relatively efficient solution which aims to use observed data to learn a functional mapping between low- and high-resolution GCMs, illustrated in Figure~\ref{fig:lr-hr}. Uncertainty in GCMs is exacerbated by both observational data and parameters in the functional mapping, motivating a probabilistic approach. 

GCMs through the Fifth Coupled Model Intercomparison Project (CMIP5) provides scientist with valuable data to study the effects of climate change under varying greenhouse gas emission scenarios~\cite{taylor2012overview}. GCMs are complex non-linear dynamical systems that model physical processes governing the atmosphere up to the year 2200 (some to 2300).  GCMs are gridded datasets with spatial resolutions around 100km and contain a range of variables including temperature, precipitation, wind, and pressure at multiple pressure levels above the earth's surface. More than 20 research groups around the world contributed to CMIP5 by developing their own models and encoding their understanding of the climate system. Within CMIP5, each GCM is simulated under three or four emission scenarios and multiple initial conditions. This suite of climate model simulations are then used to get probabilistic forecasts of variables of interest, such as precipitation and temperature extremes~\cite{semenov2010use}. While the suite of models gives us the tools to study large scale climate trends, localized projections are required for adaptation. 

Many statistical models have been explored for downscaling, from bias correction spatial disaggregation (BCSD)~\cite{cannon2002downscaling} and automated statistical downscaling (ASD)~\cite{hessami2008automated} to neural networks~\cite{taylor2000quantile} and nearest neighbor models~\cite{hidalgo2008}. Multiple studies have compared different sets of statistical downscaling approaches on various climate variables and varying temporal and spatial scales showing that no approach consistently outperforms the others~\cite{Burger2012,gutmann2014intercomparison,vandal2017intercomparison}. Recently, Vandal et al. presented improved results with an alternative approach to downscaling by representing the data as "images" and adapting a deep learning based super-resolution model called DeepSD~\cite{vandal2017deepsd}. DeepSD showed superior performance in downscaling daily precipitation in the contiguous United States (CONUS) when compared to ASD and BCSD. 

Even though uncertainty is crucial in statistical downscaling, it is rarely considered in downscaling studies. For instance, all the downscaled climate projections used in the latest US National Climate Assessment report (CSSR), produced on the NASA Earth Exchange, come with no uncertainty estimates. Though widely used in climate impact assessments, a recurrent complaint from the users is a lack of uncertainty characterization in these projections. What users often request are estimates of geographic and seasonal uncertainties such that the adaptation decisions can be made with robust knowledge~\cite{wuebbles2017us}. Khan et al. presented one study that assessed monthly uncertainty from confidence based intervals of daily predictions~\cite{khan2006uncertainty}. However, this approach only quantifies epistemic uncertainty and therefore cannot estimate a full probability distribution. To the best of the authors' knowledge, no studies have modeled aleatoric (heteroscedastic) uncertainty in statistical downscaling, presenting a limitation to adaptation. 

\subsection{Climate Data} \label{sec:data}
A wide variety of data sources exists for studying the earth's climate, from satellite and observations to climate models. Above we discussed some of the complexities and uncertainty associated with ensembles of GCMs as well as their corresponding storage and computational requirements. While the end goal is to statistically downscale GCMs, we must first learn a statistical function to apply a low- to high-resolution mapping. Fortunately, one can use observed datasets that are widely available and directly transfer the trained model to GCMs. Such observation datasets stem from gauges, satellite imagery, and radar systems. In downscaling, one typically will use either in-situ gauge estimates or a gridded data product. As we wish to obtain a complete high-resolution GCM, a gridded data product is required. Such gridded-data products are generally referred to as reanalysis datasets, which use a combination of data sources with physical characteristics aggregated to a well estimated data source. For simplicity, the remainder of this paper we will refer to reanalysis datasets as observations.

In SD, it is important for our dataset to have high spatial resolution at a daily time temporal scale spanning as many years as possible. Given these constraints, we choose to use precipitation from the Prism dataset made available by Oregon State University with a 4km spatial resolution at a daily temporal scale~\cite{daly2008physiographically}. The underlying data in Prism is estimated from a combination of gauges measuring many climate variables and topographical information. To train our model, the data is upscaled from 4km to the desired low-resolution. For example, to train a neural network to downscale from 64km to 16km, we upscale Prism to 16km and 64km and learn the mapping between the two (see Figure~\ref{fig:lr-hr}). 

For the reader, it may be useful to think about this dataset as an image where precipitation is a channel analogous to traditional RGB channels. Similarly, more variables can be added to our dataset which therefore increases the number of channels. However, it is important to be aware that the underlying spatio-temporal dynamics in the chaotic climate system makes this dataset more complex than images. In our experiments with DeepSD, we included an elevation from the Global 30 Arc-Second Elevation Data Set (GTOPO30) provided by the USGS.

\section{Background}

\subsection{DeepSD}
\begin{figure}[h]
    \centering
    \includegraphics[width=0.5\textwidth]{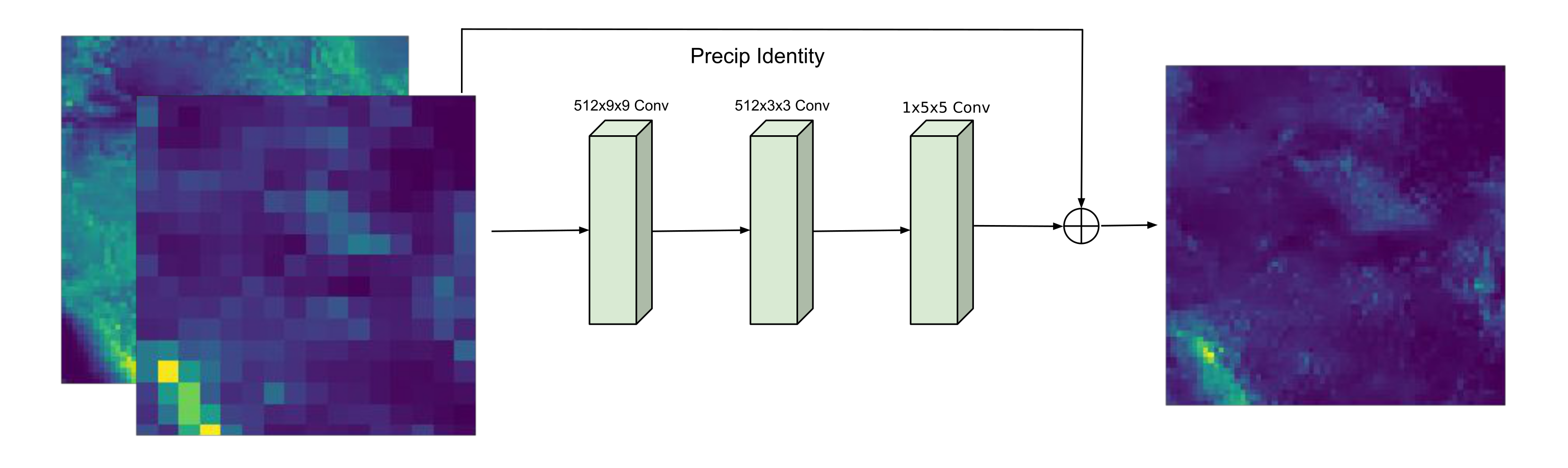}
    \caption{Residual SRCNN Architecture used for DeepSD with a skip connection between precipitation and the output layer.}
    \label{fig:deepsd-resid}
\end{figure}

The statistical downscaling approach taken by DeepSD differs from more traditional approaches, which generally do not capture spatial dependencies in both the input and output. For example Automated Statistical Downscaling (ASD)~\cite{hessami2008automated} learns regression models from low-resolution to each high-resolution point independently, failing to preserve spatial dependencies in the output and requiring substantial computational resources to learn thousands of regression models. In contrast, DeepSD represents the data as low- and high-resolution image pairs and adapts super-resolution convolutional neural networks (SRCNN)~\cite{dong2014learning} by including high-resolution auxiliary variables, such as elevation, to correct for biases. These auxiliary variables allows one to use a single trained neural network within the training domain. This super-resolution problem is essentially a pixel-wise regression such that $\Yb = F(\Xb;\Theta)$ where $\Yb$ is high-resolution with input $\Xb = [\Xb_{lr}, \Xb_{aux}]$ and $F$ a convolutional neural network parameterized by $\Theta$. $F$ can then be learned by optimizing the loss function:
 
\begin{equation} \label{eq:srcnn-loss}
    \mathcal{L}= \dfrac{1}{2N} \sum_{i \in S} \Vert F(\mathbf{X}_i;\Theta) - \mathbf{Y}_i \Vert_2^2
\end{equation}

\noindent where $S$ is a subset $n$ examples. Based on recent state-of-the-art results in super-resolution~\cite{kim2016accurate,ledig2016photo}, we modify the SRCNN architecture to include a residual connection between the precipitation input channel and output layer, as shown in Figure~\ref{fig:deepsd-resid}. 

As discussed above, the resolution enhancement of 8x or more needed in statistical downscaling is much greater than the 2-4x enhancements used for images. DeepSD uses stacked SRCNNs, each improving resolution by 2x allowing the model to capture regional and local weather patterns, depending on the level. For instance, to downscale from 100km to 12.5km, DeepSD first trains models independently (or with transfer learning) to downscale from 100km to 50km, 50km to 25km, and 25km to 12.5km. During inference, these models are simply stacked on each other where the output of one plus the next corresponding auxiliary variables are inputs to the next. In the case of downscaling precipitation, inputs may include LR precipitation and HR elevation to predict HR precipitation. In this work, we focus on uncertainty quantification for a single stacked network which can then be translated to stacking multiple Bayesian neural networks. 

\subsection{Bayesian Deep Learning}

In the early 1990's Mackay~\cite{mackay1992practical} introduced a Bayesian neural networks (BNNs) by replacing deterministic weights with distributions. However, as is common with many Bayesian modeling problems, direct inference on BNNs is intractable for networks of more than a one or two hidden layers. Many studies have attempted to reduce the computational requirements using various approximations~\cite{graves2011practical,barber1998ensemble,hinton1993keeping}. Most recently, Gal and Ghahramani presented a practical variational approach to approximate the posterior distribution in deep neural networks using dropout and monte carlo sampling~\cite{gal2016dropout,gal2016uncertainty}. Kendall and Gal then followed this work for computer vision applications to include both aleatoric and epistemic uncertainties in a single model ~\cite{kendall2017b}. 

 To begin, we define weights of our neural network as $\w = \{ \Wb_1,\Wb_2,...,\Wb_L\}$ such that $\Wb \sim \mathcal{N}(0,I)$ and $L$ being the number of layers in our network. Given random outputs of a BNN denoted by $f^{\w}(\xb)$, the likelihood can be written as $p(\yb|f^{\w}(\xb))$. 
 Then, given data $\Xb$ and $\Yb$, as defined above, we infer the posterior $p(\w | \Xb, \Yb)$ to find a distribution of parameters that best describe the data. For a regression task assuming a predictive Gaussian posterior, 
 $p(\yb|f^{\w}(\xb)) = \mathcal{N}(\hat{\yb}, \hat{\sigma}^2)$ with random outputs: 
 $$[ \hat{\yb}, \hat{\sigma}^2 ] = f^{\w}(\xb).$$ 

Applying variational inference to the weights, we can define an approximate and tractable distribution 
$q_{\Theta}(\w) = \prod_{l=1}^L q_{\M_l}(\Wb_l)$ 
where 
$q_{\M_l}(\Wb_l) = \M_l \times \text{diag}\big[\text{Bernoulli} (1-p_l)^{K_l}\big]$ 
parameterized by 
$\Theta_l = \{\M_l, p_l\}$ 
containing the weight mean of shape 
$K_l \times K_{l+1}$, $K_l$ being the number of hidden units in layer $l$,
and dropout probability $p_l$. Following, we aim to minimize the Kullback-Leibler (KL) divergence between $q_{\Theta}(\w)$ to the true posterior, $p(\w | \Xb, \Yb)$.  The optimization objective of the variational interpretation can be written as~\cite{gal2016dropout}:

\begin{align}
    \label{eq:variational-dropout}
    \mathcal{\hat{L}}(\Theta) &= -\dfrac{1}{M} \sum_{i \in S} \text{log} p(\yb_i|f^{\w}(\xb_i)) + \dfrac{1}{N} \text{KL}(q_{\Theta}(\w)|| p(\w)) \\
    &= \mathcal{\hat{L}}_x(\Theta) + \dfrac{1}{N} \text{KL}(q_{\Theta}(\w)|| p(\w))
\end{align}

\noindent where $S$ is a set of $M$ data points. To obtain well calibrated uncertainty estimates, it is crucial to select a well estimated $p_l$. Rather than setting $p_l$ to be constant, we can learn it using a concrete distribution prior which gives us a continuous approximation of the Bernoulli distribution~\cite{gal2017concrete}. As presented by Gal et al., the KL divergence term is then written as:

\begin{align}
    \label{eq:concrete-kl}
    \text{KL}(q_{\Theta}(\w)|| p(\w)) &= \sum_{l=1}^L \text{KL}(q_{\M_l}(\Wb_l)||p(\Wb_l)) \\
    \text{KL}(q_{\M_l}(\Wb)||p(\Wb)) &\propto \dfrac{l^2 (1-p_l)}{2} ||\M_l|| - K_l \mathcal{H}(p_l)
\end{align}
\noindent where 
\begin{equation}
    \label{eq:concrete-entropy}
    \mathcal{H}(p) \:= -p \text{ log } p - (1-p) \text{ log } (1-p)
\end{equation}
\noindent is the entropy of a Bernoulli random variable with probability $p$.  We note that given this entropy term, the learning dropout probability cannot exceed $0.5$, a desired effect. For brevity, we encourage the reader to refer to~\cite{gal2017concrete} for the concrete dropout optimization. In the remainder of this paper, we will use this concrete dropout formulation within all presented models.


\section{Bayesian Deep Learning for Skewed Distributions}

In this section we describe three candidate Bayesian deep learning models to quantify uncertainty in super-resolution based downscaling. We begin by formalizing the use of BDL within the SRCNN architecture assuming a normal predictive distribution, identical to the pixel-wise depth regression in~\cite{kendall2017b}. This approach is further extended to a discrete-continuous model that conditions the amount of precipitation given an occurrence of precipitation. This leverages the domain knowledge that the vast majority of data samples are non-rainy days which are easy to predict and contain little information for the regression. Such a technique was used by Sloughter el al. using a discrete-continuous gamma distribution~\cite{sloughter2007probabilistic}. Lastly, we show that a lognormal distribution can be applied directly in BDL and derive its corresponding log-likelihood loss and unbiased parameter estimates. 


\subsection{Gaussian Likelihood}

Super-resolution is an ill-posed pixel-wise regression problem such that BDL can be directly applied, as Kendall and Gal showed for predicting depth in computer vision~\cite{kendall2017b}. As discussed in previous sections, it is crucial to capture both aleatoric and epistemic uncertainties in downscaling. As shown in section 3.1 of \cite{kendall2017b}, we must measure the aleatoric uncertainty by estimating the variance, $\sigma^2$, in the predictive posterior while also sampling weights via dropout from the approximate posterior, $\widehat{\Wb} \sim q_{\Theta}(\Wb)$. As before, we defined our Bayesian convolutional neural network $\mathbf{f}$:
\begin{equation}
    \label{eq:srcnn-bdl}
    [\hat{\yb}, \hat{\sigma}^2] = \mathbf{f}^{\widehat{\Wb}}(\Xb).
\end{equation}
and make the assumption that $\Yb \sim \mathcal{N}(\hat{\yb}, \hat{\sigma}^2)$.  The Gaussian log-likelihood can be written as:
\begin{equation}
    \label{eq:normal-ll}
    \mathcal{L}_x(\Theta) = \dfrac{1}{2D} \sum_i \hat{\sigma}_i^{-2}||\yb_i - \hat{\yb}_i||^2 + \dfrac{1}{2} \text{log } \hat{\sigma}_i^{2}
\end{equation}
where pixel $i$ in $\yb$ corresponds to input $\xb$ and $D$ being the number of output pixels. The KL term is identical to that in Equation~\ref{eq:concrete-kl}. Given this formulation, $\hat{\sigma}_i$, the variance for pixel $i$ is implicitly learned from the data without the need for uncertainty labels. We also note that during training the substiution $s_i := \text{log }\hat{\sigma}_i^2$ is used for stable learning using the Adam Optimization algorithm~\cite{kingma2014adam}, a first-order gradient based optimization of stochastic objective functions. 

Unbiased estimates of the first two moments can the be obtained with $T$ Monte Carlo samples, $\{ \hat{\yb}_t, \hat{\sigma}_i^2 \}$, from $\mathbf{f}^{\widehat{\Wb}}(\xb)$ with masked weights $\widehat{\Wb_t} \sim q(\Wb)$:
\begin{align}
    \label{eq:normal-moments}
    \text{E}[\Yb] &\approx \dfrac{1}{T} \sum_{t=1}^T \hat{\yb}_t  \\
    \text{Var}[\Yb] &\approx \dfrac{1}{T} \sum_{t=1}^T \hat{\mu}_t^2 - \dfrac{1}{T} \sum_{t=1}^T \hat{\sigma}^2_t + \Big( \dfrac{1}{T} \sum_{t=1}^T \hat{\mu}_t \Big)^2.
\end{align}
These first two moments provide all the necessary information to easily obtain prediction intervals with both aleatoric and epistemic uncertainties. For further details, we encourage the reader to refer to~\cite{kendall2017b}.

\subsection{Discrete-Continuous Gaussian Likelihood}

Rather than assuming a simple Gaussian distribution for all output variables, which may be heavily biased from many non-rainy days in our dataset, we can condition the model to predict whether rain occurred or not. The BNN is now formulated such that the mean, variance, and probability of precipitation are sampled respectively from $\mathbf{f}$: 

\begin{align}
    \label{eq:mixsrcnn-output}
    [\hat{\yb}, \hat{\sigma}^2, \hat{\phi}] &= \mathbf{f}^{\widehat{\Wb}}(\Xb) \\
    \hat{p} &= \text{Sigmoid}(\hat{\phi}).
\end{align}

\noindent Splitting the distribution into discrete and continuous parts gives us:
\begin{equation}
    \label{eq:normal-peicewise}
    p\big(\yb | f^{\w}(\xb)\big) = \left \{ 
    \begin{array}{ll}
        (1-\hat{p}) & \quad \yb = 0 \\
        \hat{p} \cdot \mathcal{N}\big(\yb; \hat{\yb}, \hat{\sigma}^2\big) & \quad \yb > 0
    \end{array}
    \right.
\end{equation}

\noindent Plugging this in to~\ref{eq:variational-dropout} and dropping the constants gives us the loss function (for brevity, we ignore the KL term which is identical to Equation~\ref{eq:concrete-kl}):
\begin{equation}
    \label{eq:normalcond-loss}
    \begin{split}
        \mathcal{L}_x(\Theta) &= -\dfrac{1}{D} \sum_{i} \text{log}\Big( \mathbbm{1}_{\yb_i>0} \cdot \hat{p}_i \cdot \mathcal{N}\big(\yb_i; \hat{\yb}_i, \hat{\sigma}_i^2\big) + \mathbbm{1}_{\yb_i=0} \cdot (1-\hat{p}_i) \Big) \\
        &= -\dfrac{1}{D} \sum_{i,\yb_i>0} \Big( \text{log } \hat{p}_i + \text{log } \mathcal{N}\big(\yb_i; \hat{\yb}_i, \hat{\sigma}_i^2\big)  \Big) \\ &\hspace{4em} - \dfrac{1}{D} \sum_{i,\yb_i=0} \text{log}(1-\hat{p}_i) \\
        &= \dfrac{1}{D} \sum_{i} \big( \mathbbm{1}_{\yb_i>0} \cdot \hat{p}_i + (1-\mathbbm{1}_{\yb_i>0}) \cdot (1-\hat{p}_i) \big) \\ 
        &\hspace{4em} -\dfrac{1}{2D} \sum_{i,\yb_i>0} \hat{\sigma}^{-2}_i || \yb_i - \hat{\yb}_i ||^2 + \text{log }\sigma_i^2 
    \end{split}
\end{equation}
where the first term is the cross entropy of a rainy day and the second term is the conditional Gaussian loss. Furthermore, we can write the unbiased estimates of the first two moments as:

\begin{align}
    \label{eq:normalcond-moments}
    \text{E}[\Yb] &\approx \dfrac{1}{T} \sum_{t=1}^T \hat{p}_t \hat{\yb}_t  \\
    \text{Var}[\Yb] &\approx \dfrac{1}{T} \sum_{t=1}^T \hat{p}_t^2 \big(\hat{\yb}_t^2 + \hat{\sigma}^2_t \big) - \Big( \dfrac{1}{T} \sum_{t=1}^T \hat{p}_t \hat{\mu}_t \Big)^2.
\end{align}

\subsection{Discrete-Continuous Lognormal Likelihood}

Precipitation events, especially extremes, are known to follow fat-tailed distributions, such as lognormal and Gamma distributions~\cite{sloughter2007probabilistic,cho2004comparison}. For this reason, as above, we aim to model precipitation using a discrete-continuous lognormal distribution. It should be noted that the lognormal distribution is undefined at 0 so a conditional is required for downscaling precipitation. To do this, we slightly modify our BNN: 
\begin{align}
    \label{eq:lognormal-output}
    [\hat{\mu}, \hat{\sigma}^2, \hat{\phi}] &= \mathbf{f}^{\widehat{\Wb}}(\Xb) \\
    \hat{p} &= \text{Sigmoid}(\hat{\phi}).
\end{align}
where $\hat{\mu}$ and $\hat{\sigma}$ are sampled parameters of the lognormal distribution. Following the same steps as above, we can define a piece-wise probability density function: 
\begin{equation}
    \label{eq:lognormal-peicewise}
    p\big(\yb | f^{\w}(\xb)\big) = \left \{ 
    \begin{array}{ll}
        (1-\hat{p}) & \quad \yb = 0 \\
        \hat{p}  \cdot \dfrac{1}{\yb \hat{\sigma} \sqrt{2 \pi}} \text{exp}\Big( - \dfrac{(\text{log}(\yb) - \hat{\mu})^2}{2\hat{\sigma}^2} \Big) & \quad \yb > 0
    \end{array}
    \right.
\end{equation}
This gives us the modified log-likelihood objective:
\begin{equation}
    \label{eq:lognormalcond-loss}
    \begin{split}
        \mathcal{L}_x(\Theta) &= \dfrac{1}{D} \sum_{i} \big( \mathbbm{1}_{\yb_i>0}\cdot \hat{p}_i + (1-\mathbbm{1}_{\yb_i>0}) \cdot (1-\hat{p}_i) \big) \\ 
        &\hspace{4em} -\dfrac{1}{2D} \sum_{i,\yb_i>0} \hat{\sigma}^{-2}_i || \text{log }\yb_i - \hat{\mu}_i ||^2 + \text{log }\sigma_i^2 
    \end{split}
\end{equation}
In practice, we optimize $\hat{s} := \text{exp}(\hat{\sigma})$ for numerical stability. And lastly, the first two moments are derived as:
\begin{align}
    \label{eq:lognormalcond-moments}
    \text{E}[\yb] &\approx \dfrac{1}{T} \sum_{t=1}^T \hat{p}_t \text{exp}(\hat{\mu} + \dfrac{1}{2}\hat{\sigma}^2) \\
    \text{Var}[\Yb] &\approx \dfrac{1}{T} \sum_{t=1}^T \hat{p}_t^2  \text{exp}(2\hat{\mu} + 2\hat{\sigma}^2)
\end{align}
Given these first two moments, we can derive unbiased estimates of $\mu$ and $\sigma$:
\begin{align}
    \label{eq:lognormal-params}
    \hat{\sigma} &= \text{log} \Big( 1 + \dfrac{1}{2} \sqrt{\dfrac{4\text{Var}[\Yb]}{\text{E}[\yb]^2} + 1}\Big) \\
    \hat{\mu} &= \text{E}[\yb] - \dfrac{\hat{\sigma}^2}{2}
\end{align}
that can be used to compute pixel-wise probabilistic estimates. In the next section, we will apply each of the three methods to downscaling precipitation, compare their accuracies, and study their uncertainties. 

\section{Precipitation Downscaling}

For our experimentation, we define our problem to downscale precipitation from 64km to 16km, a 4x resolution enhancement in a single SRCNN network. We begin with precipitation from the PRISM dataset, as presented in Section~\ref{sec:data}, at 4km which is then upscaled to 16km using bilinear interpolation. This 16km dataset are our labels and are further upscaled to 64km, generating training inputs. Furthermore, we use elevation from the Global 30 Arc-Second Elevation Datset (GTOPO30) provided by the USGS as an auxilary variable, also upscaled to 16km. In the end, our dataset is made up of precipitation at 64km and elevation at 16km as inputs where precipitation at 16km are the labels. In the discrete-continuous models, precipitation >0.5mm is considered a rainy day. Precipitation measured in millimeters (mm) is scaled by $1/100$ for training when optimizing the Gaussian models. Elevation is normalized with the overall mean and variance. The training data is taken from years 1980 to 2005 and the test set from 2006 to 2015. Sub-images selected of size 64x64 with stride 48 are used for generating training examples.

Our super-resolution architecture is defined with two hidden layers of 512 kernels using kernel sizes 9, 3, and 5 (see Figure~\ref{fig:deepsd-resid}).  The model is trained for $3 \times 10^6$ iterations using a learning rate of $10^{-4}$ and a batch size of 10. Three models are optimized using each of the three log-likelihood loss's defined above, Gaussian distribution as well as discrete-continuous Gaussian and lognormal distributions conditioned on a rainy day. 50 Monte Carlo passes during inference are used to measure the first two moments which then estimates the given predictive distribution's parameters.

\begin{figure}
    \centering
    \includegraphics[width=0.5\textwidth]{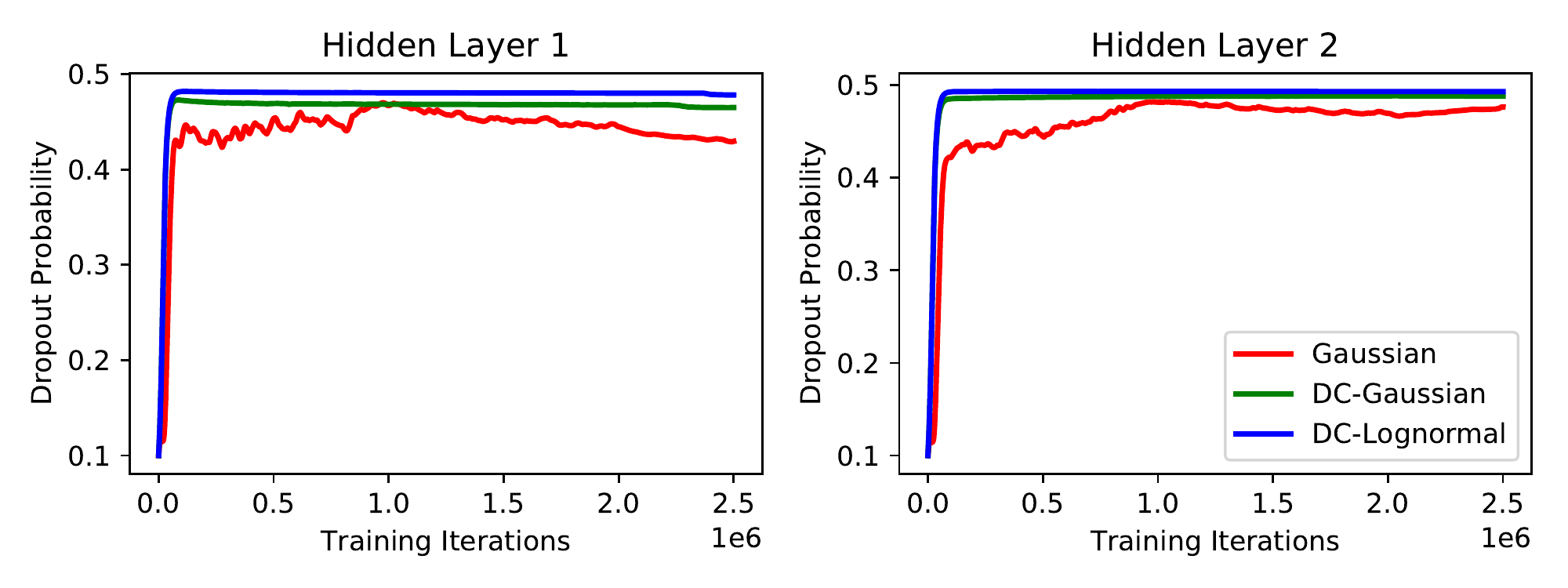}
    \caption{Dropout probabilities learned using Concrete Dropout for both hidden layers.}
    \label{fig:dropout-rates}
\end{figure}

Concrete dropout is used to optimize the dropout probability with parameters $\tau$=1e-5 and prior length scale as $l=1$ to improve uncertainly calibration performance~\cite{gal2017concrete}. For a pixel-wise regression the number of samples $N$ is set as $\text{Days} \times \text{Height} \times \text{Width}$. These parameters were found to provide a good trade-off between likelihood and regularization loss terms. As shown in Figure~\ref{fig:dropout-rates}, dropout rates for each model and hidden layer are close to 0.5, the largest possible dropout rate. We find that the Gaussian distribution has difficulty converging to a dropout rate while the discrete-continuous models quickly stabilize. Furthermore, the lognormal distribution learns the largest dropout rate, suggesting a less complex model. 

\begin{figure*}[t]
    \centering
    \includegraphics[width=\textwidth]{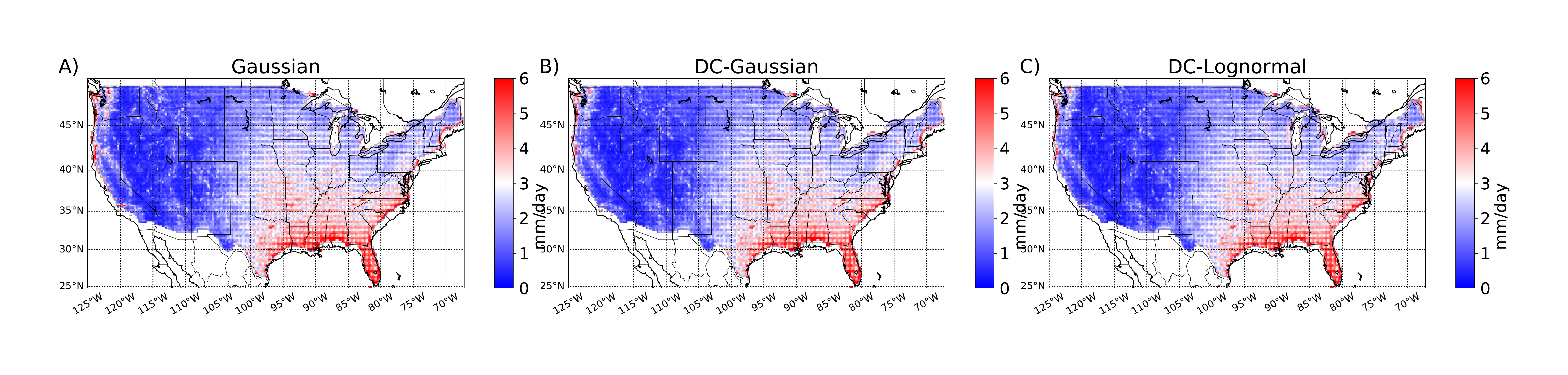}
    \caption{Daily Root Mean Square Error (RMSE) computed at each location for years 2006 to 2015 (test set) in CONUS. A) Gaussian, B) Conditional-Gaussian, and C) Conditional-Lognormal. Red corresponds to high RMSE while blue corresponds to low RMSE.}
    \label{fig:rmse-maps}
\end{figure*}

Validation is an important task for choosing a highly predictive and well calibrated downscaling model. In our experiments, we study each model's ability to predict daily precipitation, calibration of uncertainty, and width of uncertainty intervals. For reproducibility, we provide the codes for training and testing on github (\url{https://github.com/tjvandal/discrete-continuous-bdl}).

\subsection{Predictive Ability}

\begin{table*}
    \centering
    \begin{tabular}{l|r|r|r|r}
        \toprule
        {} &  Bias &  RMSE & R20 Error & SDII Error \\
        \midrule
        Gaussian     &      -0.11 $\pm$      0.34 &       2.14 $\pm$      1.31 &     -0.73 $\pm$     1.94 &      -0.83 $\pm$      0.93 \\
        DC-Gaussian  &      -0.11 $\pm$      0.30 &       2.07 $\pm$      1.28 &     -0.61 $\pm$     1.67 &      \textbf{-0.21 $\pm$      0.78} \\
        DC-Lognormal &      \textbf{-0.02 $\pm$      0.30} &       \textbf{2.05 $\pm$      1.27} &     \textbf{-0.36 $\pm$     1.63} &      -0.28 $\pm$      0.81 \\
        
        \bottomrule
    \end{tabular}
    \caption{Predictive accuracy statistics computed pixel-wise and aggregated. Daily intensity index (SDII) and yearly precipitation events greater than 20mm (R20) measure each model's ability to capture precipitation extremes. R20-Err and SDII-Err measures the difference between observed indicies and predicted indicies (closer to 0 is better).}
    \label{tab:1}
\end{table*}

We begin by comparing each model's ability to predict the ground truth observations. Root Mean Square Error (RMSE) and bias are compared to understand the average daily effects of downscaling. To analyze extremes, we select two metrics from Climdex (\url{http://www.clim-dex.org}) which provides a suite of extreme precipitation indices and is often used for evaluating downscaling models~\cite{burger2012downscaling,vandal2017intercomparison}:
\begin{enumerate}
    \item R20 - Very heavy wet days $\geq$ 20mm
    \item SDII - Daily intensity index = (Annual total) / (precip days $\geq$ 0.5 mm).
\end{enumerate}
In our analysis, we compute each index for the test set as well as observations. Then the difference between the predicted indices and observed indices are computed, ie. (SDII$_{\text{model}}$ - SDII$_{\text{obs}}$). These results can be seen in Table~\ref{tab:1}. We see a clear trend of the DC models performing better than a regular Gaussian distribution on all computed metrics. In particular, DC-Lognormal shows the lowest Bias, RMSE, and R20 error while DC-Gaussian has slightly higher errors but performs marginally better at estimating the SDII index. Furthermore, we study the predictability over space in Figure~\ref{fig:rmse-maps} by computing the pixel-wise RMSEs. Each model performs well in the mid-west and worse in the southeast, a region with large numbers of convective precipitation events. 


We see that the DC models, DC-Lognormal in particular, have lower bias than a regular Gaussian distribution. Similarly for RMSE, DC models, lead by a DC-Gaussian, have the lowest errors. Looking more closely, we see improved performance along the coasts which are generally challenging to estimate. The convolutional operation with a 5x5 kernel in the last layer reconstructs the image using a linear combination of nearby points acting as a smoothing operation. However, when this is applied to the conditional distributions, the gradient along this edge can be increased by predicting high and low probabilities of precipitation in a close neighborhood. This insight is particularly important when applied to coastal cities. 

\begin{figure}[h]
    \centering
    \includegraphics[width=0.4\textwidth]{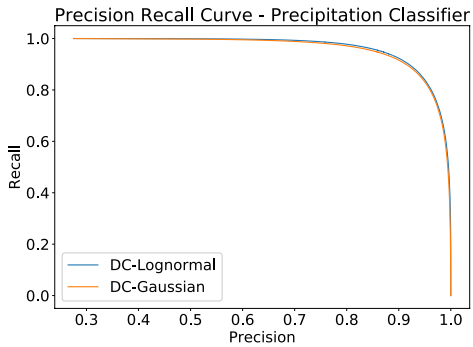}
    \caption{Precision recall curve of classifying rainy days in conditional models.}
    \label{fig:pr-curve}
\end{figure}

Lastly, we look at each conditional model's ability to classify precipitous days with precision recall curves (Figure~\ref{fig:pr-curve}). We see that recall does not begin to decrease until a precision of 0.8 which indicates very strong classification performance. It was assumed that classification of precipitation would be easy for such a dataset. 

\begin{figure*}[t]
    \centering
    \includegraphics[width=\textwidth]{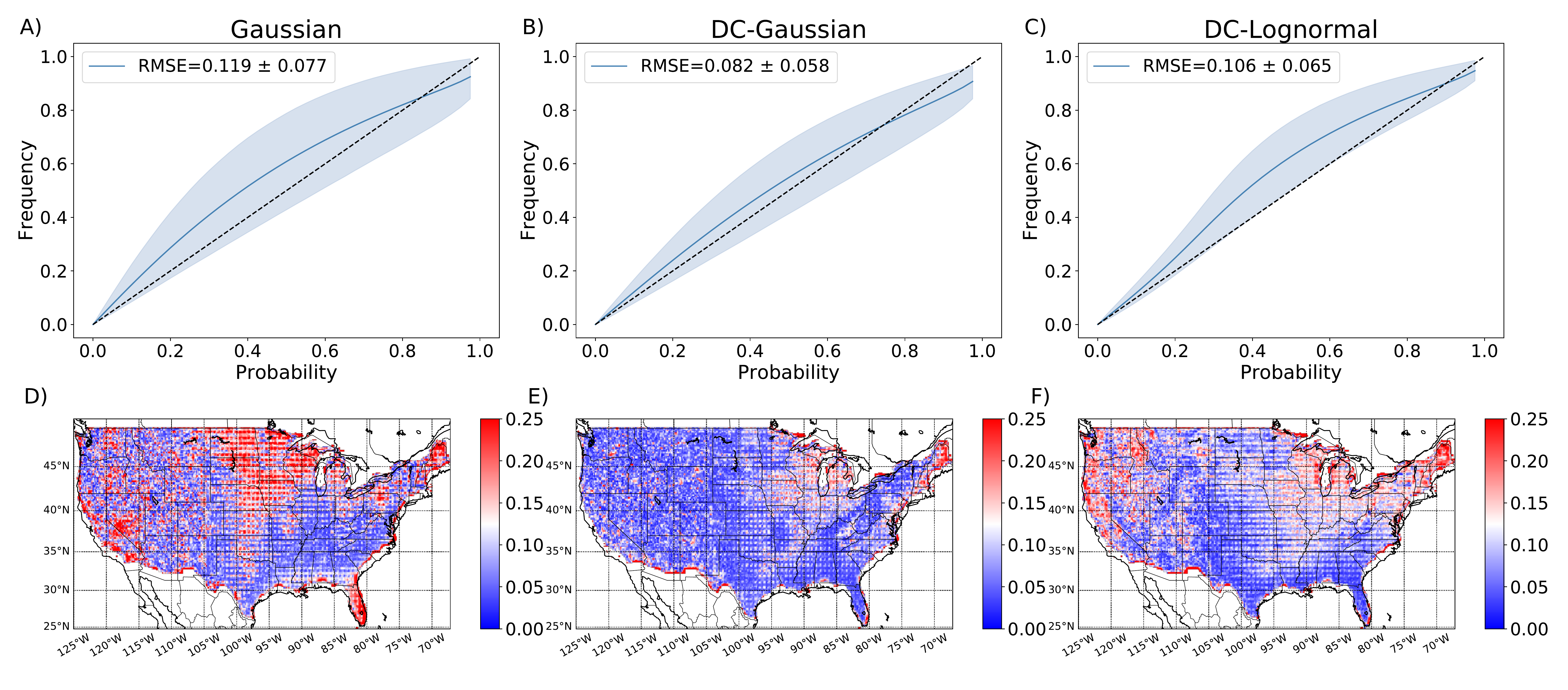}
    \caption{Calibration is computed as the frequency of predictions within a given probability range. This probability is varied on the x-axis with the corresponding frequency on the y-axis. Columns represent each model Gaussian, DC-Gaussian and Lognormal. Calibration plots on the first row compute per pixel with the shaded area representing the 80\% confidence interval of calibration. The second row depicts calibration root mean square error (RMSE) per location.}
    \label{fig:uq-calibration}
\end{figure*}

\begin{figure*}[t]
    \centering
    \includegraphics[width=\textwidth]{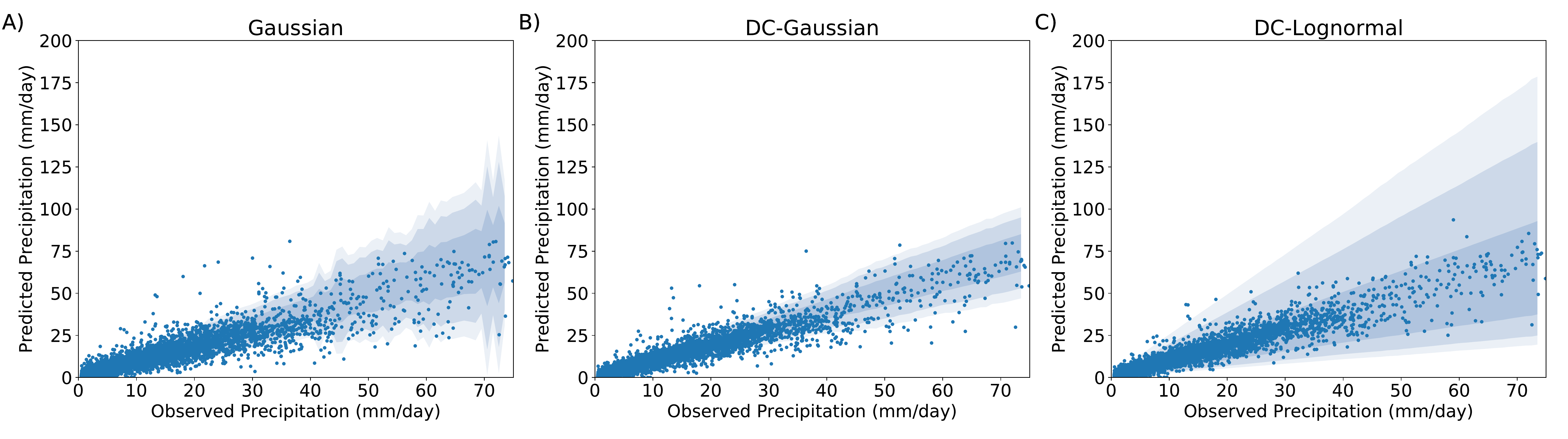}
    \caption{Uncertainty widths based on quantiles from their predictive distributions. The points are observations versus the expected value. The bands correspond to 50\%, 80\%, and 90\% predictive intervals.}
    \label{fig:uq-intervals}
\end{figure*}

\subsection{Uncertainty Quantification}

The remainder of our analysis focuses on each model's performance in estimating well calibrated uncertainty quantification. We limit our analysis of uncertainty to only days with precipitation ($\geq$ 0.5mm) as uncertainty on non-rainy days is not of interest. The calibration metric used computes the frequency of observations occurring within a varying predicted probability range:
\begin{equation}
    c(z) = \dfrac{1}{N} \sum_{i=1}^N \mathcal{I}_{P(y_i|f^{\omega}(x_i)) > (0.5 - z/2)} * \mathcal{I}_{P(y_i|f^{\omega}(x_i)) < (0.5 + z/2)}
\end{equation}
\noindent where $P$ is the cumulative density function of the predictive posterior and $z \in [0,1]$ defined the predictive probability range centered at 0.5. Ideally the frequency of observations will be equal to the probability. A calibration error can then be defined as:
\begin{equation}
    \text{RMSE}_{cal} = \sqrt{\dfrac{1}{K} \sum_{i=1}^{K} \large(c(i/K) - i/K\large)^2}
\end{equation}
\noindent where $K$ is the number bins. In our analysis, we use $K=100$. The calibration plots for each model can be seen in Figure~\ref{fig:uq-calibration}.

Right away we see from Figure~\ref{fig:uq-calibration} that the Gaussian distribution over-estimates uncertainty for most of the range with a wider range of variability between pixels. DC-Lognormal also overestimates uncertainty but has a lower range of variability between pixels, showing more consistent performance from location to location. Overall, DC-Gaussian shows the lowest calibration error hovering right around $x=y$ but underestimates uncertainty at the tails. Though DC-Lognormal is better calibrated at the tails, one could calibrate the tails by simply forcing the variance to explode. Taking this a step further, we present calibration RMSEs per pixel in Figure~\ref{fig:uq-calibration} (bottom row) to visualize spatial patterns of UQ. In the Gaussian model we find weakened and more variable results at high-elevations in the west and mid-west. Each of the DC models perform well, but DC-Lognormal also has areas of increased error in the west. 

In Figure~\ref{fig:uq-intervals} we aim to better understand these uncertainties for increasingly intense precipitation days. At these high rainfall days our models generally under-predict precipitation, but the Gaussian models often fail to capture these extremes. While the lognormal has wider uncertainty intervals, it is able to produce a well calibrated distribution at the extremes. Furthermore, these wide intervals indicate that the model becomes less confident with decreasing domain coverage at higher intensities. This may suggest that there exists a bias-variance trade-off between the Gaussian and Log-Normal distributions.

\section{Conclusion}

In this paper we present Bayesian Deep Learning approaches incorporating discrete-continuous and skewed distributions targeted at S\&E applications. The discrete-continuous models contain both a classifier to categorize an event and conditional regressor given an event's occurrence. We derive loss functions and moments for Gaussian and lognormal DC regression models. Using precipitation as an example, we condition our model on precipitous days and predict daily precipitation on a high-resolution grid. Using the lognormal distribution, we are able to produce well-calibrated uncertainties for skewed fat-tailed distributions. To our knowledge, this is the first model for uncertainty quantification in statistical downscaling. 


Through experiments, we find that this DC approach increases predictive power and uncertainty quantification performance, reducing errors with well calibrated intervals. In addition, we find that this conditional approach improves performance at the extremes, measured by daily intensity index and number of extreme precipitation days from ClimDex. Visually, we found that the DC models perform better than a regular Gaussian on the coasts, a challenge in statistical downscaling. These edge errors appear during reconstruction when the kernel partially overlaps with the coastal edge, acting as a smoothing operation. However, the DC models reduce this smoothing by increasing the expected value's gradients. 

Overall, we find that the DC distribution approaches provides strong benefits to deep super-resolution based statistical downscaling. Furthermore, while the lognormal distribution uncertainty was slightly less calibrated, it was able to produce well understood uncertainties at the extremes. This presents a strong point, Bayesian Deep Neural Networks can well fit non-normal distributions when motivated by domain knowledge.

In the future we aim to extend this work to stacked super-resolution networks, as used in DeepSD~\cite{vandal2017deepsd}, which requires sampling of between networks. Some other extensions could be the addition of more variables, extension to other skewed distributions, and larger network architectures. Finally, incorporating these theoretical advances in uncertainty characterization, the NEX team plans to use DeepSD to produce and distribute next generation of climate projections for the upcoming congressionally mandated national climate assessment.

\begin{acks}
This work was supported by NASA Earth Exchange (NEX), \grantsponsor{}{National Science Foundation CISE Expeditions in Computing}{https://www.nsf.gov/awardsearch/showAward?AWD_ID=1029711} under grant number:~\grantnum{}{1029711}, \grantsponsor{}{National Science Foundation CyberSEES}{https://www.nsf.gov/awardsearch/showAward?AWD_ID=1442728} under grant number:~\grantnum{}{1442728}, \grantsponsor{}{National Science Foundation CRISP}{https://www.nsf.gov/awardsearch/showAward?AWD_ID=1735505} under grant number:~\grantnum{}{1735505}, and \grantsponsor{}{National Science Foundation BIGDATA}{https://www.nsf.gov/awardsearch/showAward?AWD_ID=1447587} under grant number:~\grantnum{}{1447587}. The GTOPO30 dataset was distributed by the Land Processes Distributed Active Archive Center (LP DAAC), located at USGS/EROS, Sioux Falls, SD. \url{http://lpdaac.usgs.gov}. We thank Arindam Banerjee for valuable comments.
\end{acks}

\bibliographystyle{abbrv}
\bibliography{references}

\end{document}